# A Novel Generalized Artificial Neural Network for Mining Two-Class Datasets

Wei-Chang Yeh
Integration & Collaboration Laboratory
Department of Industrial Engineering and Management Engineering
National Tsing Hua University, Hsinchu, Taiwan
yeh@ieee.org

*Abstract*— A novel general neural network (GNN) is proposed for two-class data mining in this study. In a GNN, each attribute in the dataset is treated as a node, with each pair of nodes being connected by an arc. The reliability is of each arc, which is similar to the weight in artificial neural network and must be solved using simplified swarm optimization (SSO), is constant. After the node reliability is made the transformed value of the related attribute, the approximate reliability of each GNN instance is calculated based on the proposed intelligent Monte Carlo simulation (iMCS). This approximate GNN reliability is then compared with a given threshold to predict each instance. The proposed iMCS-SSO is used to repeat the procedure and train the GNN, such that the predicted class values match the actual class values as much as possible. To evaluate the classification performance of the proposed GNN, experiments were performed on five well-known benchmark datasets. The computational results compared favorably with those obtained using support vector machines.

## 1. INTRODUCTION

Data mining is an effective methodology for examining and learning from extensive compound datasets of varying quality and has been broadly applied to numerous practical problems in medicine, the social sciences, management, and engineering (Mcculloch and Pitts 1990, Hand 2007, Clifton 2010).

Various tools for data mining have been developed. An artificial neural network (ANN) has been shown to be an adequate and straightforward data-mining method (Werbos 1975, Kriesel 2007, Feng, Ong et al.

2017). It contains organized artificial neurons and can be characterized by its architecture, training method, and activation function. Different ANNs can be implemented for diverse problem types (Werbos 1975, Kriesel 2007, Feng, Ong et al. 2017).

A novel, generalized ANN is called a general neural network (GNN). One is proposed in this work for classifying two-class datasets. In the proposed GNN, its structure is extended to a more general network without limiting its layers to the input layer, hidden layer(s), and output layer.

The proposed GNN is a feedforward ANN and utilizes a supervised backpropagation learning technique that is based on machine learning. Network reliability is the fundamental concept of the proposed GNN. During the training process, the GNN maximizes accuracy by determining the reliability of each arc, with the goal of having the predicted class values match the actual class values as much as possible.

## 2. CONSTRUCT AN UBCN

A complete network is a fully connected network; that is, all nodes are interconnected such that $m = n(n − 1)/2$, where $m$ and $n$ are the number of arcs and nodes, respectively. A uniform bidirectional complete network (UBCN) is a special complete network. Its components (arcs and nodes) are either working or failing; that is, it exhibits a binary state, with each arc and/or node having its own reliability whose value is within the range of 0 to 1.

Preprocessing the proposed GNN entails constructing a UBCN by mapping each attribute (except for the class attribute) onto a node (neuron); the relationship between each pair of attributes is denoted by the reliability of the arc that connects the two nodes that represent the attribute pair. Moreover, two extra nodes, the source and sink nodes, are added to connect all the nodes.

Each arc and node has its own reliability. The value of each arc's reliability is generated randomly and uniformly within the range [0, 1] at the beginning and updated generation by generation based on the machine-learning methods in this study. Conversely, the value of each node's reliability is equal to the normalized attribute values, which are within the range of 0 to 1.

The remainder of this study focuses on calculating the reliability of the GNN, which relies more on Monte Carlo simulation (MCS) and simplified swarm optimization (SSO).

## 3. DATA TRANSFORMATION

One part of the preprocessing of the GNN involves adjusting all values in a dataset.

### A. Transforming the Class

We developed a first version of this model, that can be applied only to two classes data sets. Let $y_i$ be the class attribute of the $i$th instance. The value of $y_i$ is transformed to 1 if the number of instances with the same class as $y_i$ is larger than the number of instances without the same class as $y_i$; otherwise, $y_i$ is set to 0.

### B. Data Normalization Based on the Correlation

The range of the reliability of each component (i.e., nodes and arcs) is from 0 to 1. Component reliability is highly correlated with network reliability (Colbourn 1987, Jin and Coit 2003, Levitin 2005, Pham 2007, Zuo, Tian et al. 2007). Hence, the preprocessing of the GNN includes adjusting the dataset so that the following property is also satisfied:

**Property 1.** A higher network reliability indicates that each component is more likely to have a greater reliability.

As discussed above, the values in the $i$th attribute $A_i$ must all be normalized and transformed, such that their values are within the range [0, 1], and must also satisfy Property 1; this is achieved by performing the following equation:

$$x_{i,j}^{\#} = \begin{cases} \dfrac{x_{i,j} - \underset{k}{Min}\{x_{i,k}\}}{\underset{k}{Max}\{x_{i,k}\} - \underset{k}{Min}\{x_{i,k}\}} & \text{if } 0 \leq r_s(\tilde{A}_i, \tilde{Y}) \\ 1 - \dfrac{x_{i,j} - \underset{k}{Min}\{x_{i,k}\}}{\underset{k}{Max}\{x_{i,k}\} - \underset{k}{Min}\{x_{i,k}\}} & \text{otherwise} \end{cases},$$

where $r_s(\tilde{A}_i, \tilde{Y})$ is the Spearman's rank-order correlation, $\tilde{A}_i$ is the ranked $A_i$, $\tilde{Y}$ is the ranked transformed class attribute, and $x_{i,j}$ and $x_{i,j}^{\#}$ are the original value and transformed value, respectively, of the $j$th attribute in the $i$th instance.

## 4. THE PROPOSED iMCS

It is an NP-hard problem to calculate the exact reliability of a network. Hence, because MCS is an efficient and commonly used tool (Diaz-Emparanza 1997, Jin and Coit 2003, Yeh 2017), it is used in the GNN to obtain the approximate network reliability $R^*$.

Basically, MCS checks whether it can successfully find a path from the source node to the sink node after generating the reliability values of the components randomly. The above procedure is simulated for a certain number, say $N_{sim}$, with the success rate being $R^*$, which is an unbiased and consistent estimator of complex network reliability (Diaz-Emparanza 1997, Jin and Coit 2003, Yeh 2017).

### 4.1 Threshold θ

After all arc reliabilities are known, the $R^*$ value of the $i$th instance can be obtained using MCS after substituting the related transformed value in the $i$th instance for the node reliability. Then, the $R^*$ value is compared with a predefined threshold θ to predict the class of the $i$th instance $y_i^*$, such that

$$y_i^* = \begin{cases} 0 & \text{if } \theta < R^* \\ 1 & \text{otherwise} \end{cases}.$$

The threshold θ is defined as the value of the ratio of class 1, based on Section IIIA.

**4.2 Reducing the Replication Number**

The total replication number of the MCS used in machine learning is

$$N_{run} \cdot N_{gen} \cdot N_{sol} \cdot N_{rec} \cdot N_{sim},$$

where $N_{run}$, $N_{gen}$, $N_{sol}$, and $N_{rec}$ are the number of runs, generations, solutions, and instances, respectively. In this study, $N_{run} = 30$, $N_{gen} = 50$, $N_{sol} = 10$, and $N_{sim} = 2000$. For example, the total replication number is $3 \cdot 10^{10}$ if $N_{rec} = 1000$.

Hence, it remains very tedious to use MCS to calculate the approximate reliability, particularly when the network reliability must be obtained for each instance in each solution of each generation. Therefore, an intelligent MCS (named "iMCS" here) is proposed to reduce the replication number and improve the efficiency of the MCS.

The following properties summarize the statistical characteristics of $R^*$ in relation to the minimal replication number under a certain error and confidence interval (Diaz-Emparanza 1997). (Its proof can be found in any textbook related to reliability.)

**Property 2:** The number $N_{sim}$ under the confidence interval $(1-\alpha)\%$ must be taken at least

$$N_{sim} \geq Z_{\alpha/2}^2 \frac{p_\varepsilon (1 - p_\varepsilon)}{\varepsilon^2}.$$

where

$\varepsilon$ = the relative error;

$p_\varepsilon$ = the probability of an estimator with $\varepsilon$;

$Z_{\alpha/2}$ = the upper $\alpha/2$ percentage point of the standard normal distribution.

Let $N_{sim}$ be separated into equal-length intervals such that the interval length is $\Delta N_{sim}$, and the left and right endpoints in the $i$th interval are $N_{sim}^{(i-1)}$ and $N_{sim}^{(i)}-1$, respectively, where $N_{sim}^{(i)}=N_{sim}^{(i-1)}+\Delta N_{sim}$ and $N_{sim}^{(0)}=0$. From Property 2, we have the following property.

**Property 3:** Let

$$\omega_i = \sum_{k=1}^{i} m_i$$

$$\Delta_i = Z_{\alpha/2}[\sqrt{\frac{p_\varepsilon(1-p_\varepsilon)}{N_{sim}^{(i)}}} - \sqrt{\frac{p_\varepsilon(1-p_\varepsilon)}{N_{sim}}}],$$

where $m_i$ is the successful number in MCS during ($N_{sim}^{(i)}-N_{sim}^{(i-1)}$) simulated times. Then,

$$N_{sim}^{(i)}(R_{Nsim}-\Delta_i) \leq \omega_i \leq N_{sim}^{(i)}(R_{Nsim}+\Delta_i).$$

**Proof:** From Property 2, we have

$$\varepsilon^2 \geq Z_{\alpha/2}^2 \frac{p_\varepsilon(1-p_\varepsilon)}{M}$$

$$\Rightarrow \varepsilon \geq Z_{\alpha/2}\sqrt{\frac{p_\varepsilon(1-p_\varepsilon)}{N_{sim}}}$$

$$\Rightarrow |R-R_{Nsim}| \geq Z_{\alpha/2}\sqrt{\frac{p_\varepsilon(1-p_\varepsilon)}{N_{sim}}}$$

$$\Rightarrow R_{Nsim} \geq R + Z_{\alpha/2}\sqrt{\frac{p_\varepsilon(1-p_\varepsilon)}{N_{sim}}}$$

$$\Rightarrow R_{Nsim}^{(i)} - R_{Nsim} \geq Z_{\alpha/2}[\sqrt{\frac{p_\varepsilon(1-p_\varepsilon)}{N_{sim}^{(i)}}} - \sqrt{\frac{p_\varepsilon(1-p_\varepsilon)}{N_{sim}}}]$$

$$\Rightarrow R_{Nsim}^{(i)} - R_{Nsim} \geq \Delta_i$$

$$\Rightarrow R_{Nsim} - \Delta_i \leq R_{Nsim}^{(i)}.$$

In the same way,

$$R_{Nsim}^{(i)} \leq R_{Nsim} + \Delta_i.$$

Hence,

$$R_{\text{Nsim}} - \Delta_i \leq R_{\text{Nsim}}^{(i)} \leq R_{\text{Nsim}} + \Delta_i$$

$$\Rightarrow N_{\text{sim}}^{(i)}(R_{\text{Nsim}} - \Delta_i) \leq \omega_i \leq N_{\text{sim}}^{(i)}(R_{\text{Nsim}} + \Delta_i).$$

□

Proceeding from Property 3, this property is the most critical in the proposed GNN. It is able to reduce the replication number intelligently:

**Property 4:** The predicted class is 0 and 1 of the related instance if

$$\omega_i \leq N_{\text{sim}}^{(i)}(\theta - \Delta_i) = \text{LB}_i$$

and

$$\text{UB}_i = N_{\text{sim}}^{(i)}(\theta + \Delta_i) \leq \omega_i,$$

respectively.

**Proof:** From the definition of $\theta$ and Property 3,

$$y_i^* = \begin{cases} 0 & \text{if } R_{\text{Nsim}} < \theta \\ 1 & \text{otherwise} \end{cases},$$

and

$$N_{\text{sim}}^{(i)}(R_{\text{Nsim}} - \Delta_i) \leq \omega_i \leq N_{\text{sim}}^{(i)}(R_{\text{Nsim}} + \Delta_i)$$

under the confidence interval $(1-\alpha)\%$. Hence, we have

$$y_i^* = \begin{cases} 0 & \text{if } \omega_i < N_{\text{sim}}^{(i)}(\theta - \Delta_i) \\ 1 & \text{if } N_{\text{sim}}^{(i)}(\theta + \Delta_i) < \omega_i \end{cases}$$

when the replication number is $N_{\text{sim}}^{(i)}$. □

Note that

1. there is no conclusion, and additional simulations are required to predict the related instance if

$$N_{\text{sim}}^{(i)}(\theta - \Delta_i) \leq \omega_i \leq N_{\text{sim}}^{(i)}(\theta + \Delta_i);$$

2. if $i$ is the last interval, $LB_i=UB_i$ because $N_{sim}^{(i)}=N_{sim}$ and $\Delta_i=0$;

3. the confidence interval $(1-\alpha)\%$ is 99%, $p_\varepsilon=0.90$, $N_{sim}=2000$, and $\Delta N_{sim}=100$, e.g., $N_{sim}^{(5)}=500$, in the proposed iMCS.

**4.3 The proposed iMCS**

Based on Property 4, the proposed iMCS is listed in the following to reduce replication numbers.

PROCEDURE iMCS($k$)

**Input:** The $k$th instance.

**Output:** The predicted class value, which is either 0 or 1, of the $k$th instance.

**STEP 0.** Let $i=1$ and $\omega_0=0$.

**STEP 1.** Let $\omega_i=\omega_{i-1}+m_i$, where $m_i$ is defined in Property 3.

**STEP 2.** If $\omega_i \leq LB_i$, then halt and return 0.

**STEP 3.** If $\omega_i \geq UB_i$, then halt and return 1.

**5. THE PROPOSED iMCS-SSO**

The simplified swarm optimization (SSO) proposed by Yeh is the simplest of all machine-learning methods (Yeh , Yeh 2012, Yeh 2013, Yeh 2017). In addition, SSO has been shown to be a very powerful tool in data mining for certain datasets (Yeh 2012, Yeh 2013, Yeh 2017). Therefore, SSO is implemented to determine the arc reliabilities in the GNN.

**5.1 SSO**

Let $N_{var}$ be the number of variables (i.e., the number of arcs in the proposed model) in each solution, $X_i=(r_{i,1}, r_{i,2}, \ldots, r_{i,Nvar})$ be the $i$th solution, $P_i=(p_{i,1}, p_{i,2}, \ldots, p_{i,Nvar})$ be the best solution among all the $i$th solutions, $G=(g_1, g_2, ..., g_{Nvar})$ be the best solution among all solutions, where $i=1, 2, ..., N_{sol}$. The new $r_{i,j}$ is obtained based on the following step function (Yeh, Yeh 2012, Yeh 2013, Yeh 2017):

$$r_{i,j} = \begin{cases} g_j & \text{if } \rho_{[0,1]} \in [0, c_g) \\ p_{i,j} & \text{if } \rho_{[0,1]} \in [c_g, c_g + c_p) \\ r_{i,j} & \text{if } \rho_{[0,1]} \in [c_g + c_p, c_g + c_p + c_w) \\ x & \text{otherwise} \end{cases}$$

where $\rho_{[0,1]}$ is a random number generated from uniform distribution from 0 to 1; $x$ is a random value generated from the feasible region of the $j$th variable; $c_g$, $c_p$, $c_w$, and $1-c_g-c_p-c_w$ are the predefined probabilities that the new updated value of $r_{i,j}$ are copied from the current value of $g_j$, $p_{i,j}$, and $r_{i,j}$, respectively. In the proposed algorithm, $c_g = 0.4$, $c_p = 0.2$, and $c_w = 0.1$.

## 5.2 Calculating the Fitness

In data mining, each solution of any instance-based machine learning method requires checking each instance one by one to calculate the ratio of the correct predicted classes, which is the fitness of the solution.

**PROCEDURE iMCS-FITNESS($X_{sol}$)**

**Input:** The solution $X_{sol}$.

**Output:** The fitness value $F(X_{sol})$ of $X_{sol}$.

**STEP 0.** Let rec=1 and $f=0$.

**STEP 1.** Let $f = f + $ iMCS(rec).

**STEP 2.** If rec<Nrec, let rec=rec+1 and go to STEP 1.

**STEP 3.** Return $f$ / $N_{rec}$.

## 5.3 The Pseudo Code for the Proposed iMCS-SSO

The pseudo code for the proposed iMCS-SSO integrates the preprocessing procedure, the updating of solutions achieved through SSO, and the calculating of the fitness of each solution (i.e., the iMCS-FITNESS) achieved by performing the proposed iMCS.

**PROCEDURE iMCS-SSO()**

**Input:** A data set.

**Output:** A classifier.

**STEP 0.** Build an UBCN for the dataset based on Section II, transform dataset based on Section III, calculate $UB_i$ and $LB_i$ for $i=1, 2, \ldots, N_{sim}/100$.

**STEP 1.** Generate solutions $X_i$ randomly, let $F(X_{sol})=$iMCS-FITNESS$(X_{sol})$, $P_{sol}=X_{sol}$, $G=X_k$, and gen=2, where $F(X_k)=$Max$\{F(X_{sol})\}$ and sol=1, 2, …, $N_{sol}$.

**STEP 2.** Let sol=1.

**STEP 3.** Update $X_{sol}$ based on the SSO.

**STEP 4.** Let $F(X_{sol})=$iMCS-FITNESS$(X_{sol})$.

**STEP 5.** If $F(X_{sol})>F(P_{sol})$, let $P_{sol}=X_{sol}$. Otherwise, go to STEP 7.

**STEP 6.** If $F(X_{sol})>F(G)$, let $G=X_{sol}$.

**STEP 7.** If sol<$N_{sol}$, let sol=sol+1 and go to STEP 3.

**STEP 8.** If gen<$N_{gen}$, let gen=gen+1 and go to STEP 2.

**STEP 9.** Halt and $G$ is the classifier we need.

## 6. EXPERIMENTAL RESULTS AND SUMMARY

To demonstrate the performance of the proposed GNN, it was tested on five famous two-class datasets: "Australian Credit Approval" (A), "breast-cancer" (B), "diabetes" (D), "fourclass" (F) (Ho and Kleinberg 1996) and "Heart Disease" (H) based on a tenfold cross-validation. In addition, the performance of the proposed GNN was compared with that of a support vector machine (SVM).

TABLE I.

The accuracy of SVM and GNN for 5 benchmark datasets.

| Datasets \ Fold | A | | B | | D | | F | | H | |
|---|---|---|---|---|---|---|---|---|---|---|
| | SVM | GNN | SVM | GNN | SVM | GNN | SVM | GNN | SVM | GNN |
| 1 | 81.666664% | **85.000000%** | 95.588234% | **97.058823%** | **76.712326%** | **76.712326%** | 80.232559% | **82.558136%** | 80.000000% | **84.000000%** |
| 2 | **92.156860%** | **92.156860%** | **100.000000%** | **100.000000%** | **76.811592%** | 75.362320% | 79.000000% | **81.000000%** | 82.758621% | **86.206894%** |
| 3 | 83.116882% | **89.610390%** | 95.588234% | **97.058823%** | **82.022469%** | 79.775284% | 75.862068% | **77.011497%** | 82.758621% | **89.655174%** |
| 4 | 92.187500% | **93.750000%** | **100.000000%** | **100.000000%** | 76.470589% | **80.882355%** | **81.818184%** | **81.818184%** | 88.235291% | **97.058823%** |
| 5 | 87.878792% | **89.393936%** | **100.000000%** | **100.000000%** | 71.232880% | **76.712326%** | 82.857140% | **84.285713%** | **100.000000%** | 95.000000% |
| 6 | 80.701752% | **82.456139%** | 94.117645% | **97.058823%** | 68.292686% | **74.390244%** | **81.914894%** | 80.851067% | 80.000000% | **90.000000%** |
| 7 | 82.278481% | **86.075951%** | 95.588234% | **98.529411%** | **82.558144%** | 81.395348% | **85.897438%** | **85.897438%** | 82.608696% | **95.652176%** |
| 8 | 85.135132% | **87.837837%** | 98.529411% | **100.000000%** | 71.428574% | **76.623375%** | 74.157303% | **77.528091%** | **89.285713%** | 85.714287% |
| 9 | 86.419754% | **88.888885%** | 94.117645% | **95.588234%** | 80.519478% | **83.116882%** | 75.581398% | **79.069771%** | 75.000000% | **83.333336%** |
| 10 | 85.185188% | **86.419754%** | 97.183098% | **97.183098%** | 78.378380% | **79.729729%** | **86.904762%** | 84.523811% | 84.615387% | 84.615387% |
| Avg. | 85.672701% | **88.158975%** | 97.071250% | **98.247721%** | 76.442712% | **78.470019%** | 80.422575% | **81.454371%** | 84.526233% | **89.123608%** |

TABLE II.
The accuracy of SVM and GNN for 5 benchmark datasets.

| Fold | A | B | D | F | H |
|------|---|---|---|---|---|
| 1    | 138.56 (6.93%) | 107.30 (5.37%) | 232.28 (11.61%) | 159.15 (7.96%) | 140.40 (7.02%) |
| 2    | 131.24 (6.56%) | 111.57 (5.58%) | 217.25 (10.86%) | 166.73 (8.34%) | 141.03 (7.05%) |
| 3    | 119.96 (6.00%) | 102.94 (5.15%) | 219.14 (10.96%) | 179.46 (8.97%) | 127.93 (6.40%) |
| 4    | 125.00 (6.25%) | 103.73 (5.19%) | 202.84 (10.14%) | 150.11 (7.51%) | 132.06 (6.60%) |
| 5    | 124.09 (6.20%) | 104.07 (5.20%) | 189.54 (9.48%)  | 158.76 (7.94%) | 145.33 (7.27%) |
| 6    | 118.19 (5.91%) | 109.80 (5.49%) | 187.72 (9.39%)  | 151.38 (7.57%) | 133.50 (6.68%) |
| 7    | 126.41 (6.32%) | 112.84 (5.64%) | 178.95 (8.95%)  | 174.40 (8.72%) | 164.20 (8.21%) |
| 8    | 126.89 (6.34%) | 104.61 (5.23%) | 208.83 (10.44%) | 162.66 (8.13%) | 135.48 (6.77%) |
| 9    | 122.67 (6.13%) | 112.79 (5.64%) | 181.08 (9.05%)  | 177.40 (8.87%) | 164.07 (8.20%) |
| 10   | 125.51 (6.28%) | 115.73 (5.79%) | 181.53 (9.08%)  | 164.17 (8.21%) | 139.49 (6.97%) |
| Avg. | 125.85 (6.29%) | 108.54 (5.43%) | 199.92 (10.00%) | 164.42 (8.22%) | 142.35 (7.12%) |

TABLE I shows the highest accuracy obtained in each fold on all five datasets for both the SVM and the proposed GNN; TABLE II lists the average simulation numbers used in the GNN after implementing the proposed iMCS-SSO.

As shown in TABLE I, the proposed GNN outperformed the SVM on each dataset with an improved accuracy of at least 1.03% and up to 4.60%. In addition, the GNN performed better than or equal to the SVM for each fold on datasets A and B. As shown in TABLE II, implementing the proposed iMCS-SSO in training the GNN achieved clearly superior outcomes on these five datasets, reducing the simulation numbers by at least 10% compared with traditional MCS.

## 7. CONCLUSIONS AND FUTURE WORK

Data mining is a critical means of providing useful information in the contemporary world. In this pilot study, the proposed GNN is a novel, generalized ANN that is trained using the proposed iMCS-SSO. The results of the experiments on the five benchmark datasets (Section VI) show that this GNN improves accuracy and that the iMCS effectively reduces the simulation number in all cases. Given these outcomes, the proposed

GNN should be extended, with future studies applying it to multi-class datasets with more attributes, classes, and instances.

## ACKNOWLEDGMENTS

This research was supported in part by the National Science Council of Taiwan, R.O.C. under grant MOST 101-2221-E-007-079-MY3 and MOST 102-2221-E-007-086-MY3.